\documentclass[letterpaper, 10 pt, journal, twoside]{IEEEtran}  % Comment this line out if you need a4paper

\usepackage{graphicx}
\usepackage{amsmath}
\usepackage{amssymb}
\usepackage{booktabs}
\usepackage{adjustbox}
\usepackage[dvipsnames]{xcolor}
\usepackage{threeparttable} % for table footnote
\usepackage{hyperref} % for link footnote
\usepackage{subcaption}

\usepackage{harpoon}% <---

\usepackage[linesnumbered,ruled,vlined]{algorithm2e}
\SetKwInput{KwInput}{Input}                % Set the Input
\SetKwInput{KwOutput}{Output}              % set the Output

\usepackage{pstool}

\usepackage{siunitx}

\usepackage{multirow}

\usepackage[capitalize]{cleveref}
\crefname{section}{Sec.}{Secs.}
\Crefname{section}{Section}{Sections}
\Crefname{table}{Table}{Tables}
\crefname{table}{Tab.}{Tabs.}

\usepackage{tikz}
\def\checkmark{\tikz\fill[scale=0.2](0,.35) -- (.25,0) -- (1,.7) -- (.25,.15) -- cycle;} 
\newcommand{\tabincell}[2]{\begin{tabular}{@{}#1@{}}#2\end{tabular}}

\definecolor{revised}{cmyk}{0, 0.7808, 0.4429, 0.1412}

\newcommand*{\M}{\mathbf}
\newcommand*{\V}{\mathbf}

\DeclareMathOperator*{\argmin}{arg\,min}

\title{
	Indirect Point Cloud Registration: \\ Aligning Distance Fields
  using a \\ Pseudo Third Point Set}

\author{Yijun Yuan and Andreas N\"uchter% <-this % stops a space
\thanks{All authors are with Informatics VII : Robotics and Telematics, University of W{\"u}rzburg
	{\tt\small \{yijun.yuan|andreas. nuechter\}@uni-wuerzburg.de }}%
}

\begin{document}

%\markboth{IEEE Robotics and Automation Letters. Preprint Version. Accepted June, 2022}
%{Yuan \MakeLowercase{\textit{et al.}}: Indirect Point Cloud Registration} 

\maketitle

\IEEEpeerreviewmaketitle
%%%%%%%%% ABSTRACT
\begin{abstract}
In recent years, implicit functions have drawn attention in the field of 3D reconstruction and have successfully been applied with Deep Learning.
However, for incremental reconstruction, implicit function-based registrations have been rarely explored.
Inspired by the high precision of deep learning global feature registration, we propose to combine this with distance fields.
%In this paper,
We generalize the algorithm to a non-Deep Learning setting while retaining the accuracy.
%However the recent used registration with the implicit function is still the conventional point-to-implicit design. Which make it impossible to 
%
%In the Deep Learning based registration, global feature registration has been newly explored. Those SOTA methods rely on feature metric to iteratively solve the result. We observe that the framework used in the class of algorithms is able to generalized to the non-Deep Learning which has also not been explored. 
%Thus in this paper, we propose registration between distance fields algorithm based on Feature-metric Registration framework that
Our algorithm is more accurate than conventional models while, without any training, it achieves a competitive performance and faster speed, compared to Deep Learning-based registration models.
%
%The algorithm propose to introduce a third smaller random point set to register points. During registration, the two frames to align are fixed, we merely move the random generated small set of points.
%In experiments, we observe that the algorithm works faster and more accurately than the conventional methods.
%Without any training, our model obtained a competitive performance.
%FIXED: one of the above sentences suffices.
The implementation is available on github\footnote{\url{https://github.com/Jarrome/IFR}} for the research community.
\end{abstract}
\begin{IEEEkeywords}
	Localization; Mapping; SLAM
\end{IEEEkeywords}
%%%%%%%%% BODY TEXT
\section{Introduction}
\label{sec:intro}
% 1. development of Point cloud processing
\IEEEPARstart{D}{uring} the fast development of artificial intelligence, perception always has been playing an important role as it is the most fundamental component for other tasks.
In the field of perception, 3D laser scanners producing 3D point clouds are the most direct and accurate tools to perceive the geometry of the environment of a robotic system.
Reconstruction, detection, tracking, etc., are typical tasks of a perception system.

% 1.1 3D reconstruction
3D reconstruction, with its high potential for commercial applications, has recently received a lot of attention.
Major contributions in this field include methods using Truncated Signed Distance Field (TSDF) based algorithms~\cite{koch2016multi}.
% This paer would give a good cite
%Koch, P., May, S., Schmidpeter, M., Kühn, M., Pfitzner, C., Merkl, C., Koch, R., Fees, M., Martin, J., Nüchter, A.: Multi-Robot Localization and Mapping based on Signed Distance Functions. Journal of Intelligent and Robotic Systems. 83, 409--428 (2016).
By using continuous implicit functions, shapes can be represented in arbitrary resolution with high quality.  
To build reconstructions incrementally, frame-to-frame pose transformations are required.
Here, the Point Cloud Registration (PCR) algorithm plays an important role.
% 1.5. the important role of PCR 
%Point Cloud Registration (PCR), as a most elemental function in the incremental reconstruction, also plays an important role.
%Point cloud registration is one of the most elemental function in SLAM system. 
The performance of the feature extraction and association strongly affects the performance of the registration algorithm~\cite{yuan2021self}.
% DONE
% Would it be here good to cite our previous work, e.g., the MDPI paper
% 2. PCR conventional method

% point-to-implicit
For the 3D reconstruction application, the point-to-implicit technique is commonly used because it shows high utility for its capability to seamlessly couple registration into the reconstruction framework~\cite{huang2021di}.
% FIXED: cite here also the orioginal kinect fusion paper? kinect fusion is with ICP not point-to-implicit ?!
By minimizing the total energy of one point cloud on the field of a given point cloud, the transformation is optimized.
However, this class of registration requires to iteratively fit one point cloud in the distance field of the other, which is in-efficient. 
The seminal work by M. Slavcheva proposes an implicit-to-implicit algorithm and improves the accuracy of registrations~\cite{slavcheva2018signed}.
However, this method requires discretizing the field into dense voxels of distance values, which represents the function with volumes.
The advantage of a dense representation is that it provides voxel-to-voxel matched.
But the disadvantage is the inefficiency both in space and time. Thus, it is used only for small object reconstruction.
In this paper, we focus on a sparse solution for implicit-to-implicit registration.

%the proposed method is able to works on synthetic, real indoor and real outdoor point clouds.

%\begin{figure}[t!]
%	\centering
%	%\fbox{\rule{0pt}{2in} \rule{0.9\linewidth}{0pt}}
%	\begin{subfigure}{.5\textwidth}
%		\centering
%		\includegraphics[width=0.8\linewidth]{im/moon1.png}
%		%\caption{a}
%	\end{subfigure}%
%	\\
%	\begin{subfigure}{.5\textwidth}
%		\centering
%		\includegraphics[width=0.8\linewidth]{im/moon2.png}
%		%\caption{b}
%	\end{subfigure}%
%%FIXED: The colors in the caption are interesting. But do they match the ones in the fiugure? I strongly reocmment to use the linux program xfig, export he figure to an .eps-file and use \psfrag for putting the actual LaTeX font into the figure    
%	\caption{Diagram of Psudo-point set Registration. The \textcolor{BlueGreen}{$P$} and \textcolor{Orange}{$Q$} are two frames to register. \textcolor{Purple}{$S$} dots and \textcolor{YellowOrange}{$S^{'}$} dots are psudo-points. During registration, only \textcolor{YellowOrange}{$S^{'}$} is used, \textcolor{Blue}{$P$} and \textcolor{Orange}{$Q$} are fixed.}
%	\label{fig:Diagram}
%\end{figure}
\begin{figure}[t!]
	\centering
	\begin{subfigure}{.5\textwidth}
		\centering
		\psfragfig[width=.4\textwidth]{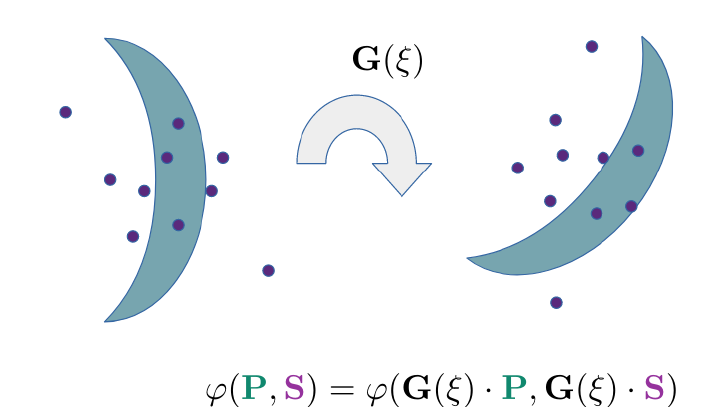}{
			\psfrag{R}{$\V G(\xi)$}
				\psfrag{F(P,S)=F(GP,GS)}{$\varphi(\textcolor{PineGreen}{\V P},\textcolor{Purple}{\V S}) = \varphi(\V G(\xi)\cdot \textcolor{PineGreen}{\V P},\V G(\xi)\cdot \textcolor{Purple}{\V S})$}
		}
		%\caption{a}
	\end{subfigure}%
	\\
	\begin{subfigure}{.5\textwidth}
		\centering
		\psfragfig[width=.48\textwidth]{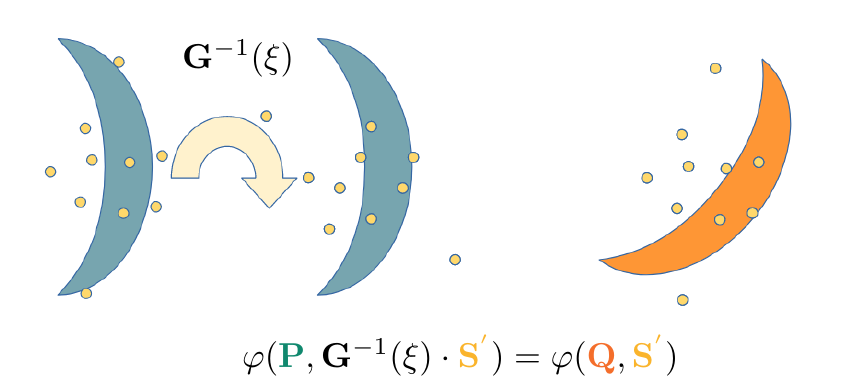}{
	\psfrag{R}{$\V G^{-1}(\xi)$}
	\psfrag{F(P,GS)=F(Q,S)}{$\varphi(\textcolor{PineGreen}{\V P},\V G^{-1}(\xi)\cdot \textcolor{Dandelion}{\V S^{'}}) = \varphi(\textcolor{Orange}{\V Q},\textcolor{Dandelion}{\V S^{'}})$}
}		
		
		%\caption{b}
	\end{subfigure}%
	%FIXED: The colors in the caption are interesting. But do they match the ones in the fiugure? I strongly reocmment to use the linux program xfig, export he figure to an .eps-file and use \psfrag for putting the actual LaTeX font into the figure    
	\caption{Diagram of Pseudo-point set registration. $\varphi$ is a transformation invariant function. 
       \textcolor{PineGreen}{$\V P$} and \textcolor{Orange}{$\V Q$} in \textbf{moon shape} are two frames to register.
       The dots \textcolor{Purple}{$\V S$} and \textcolor{Dandelion}{$\V S^{'}$} are the pseudo-points.
       While registration to solve $\V G^{-1}(\xi)$, only \textcolor{Dandelion}{$\V S^{'}$} is used, \textcolor{PineGreen}{$\V P$} and \textcolor{Orange}{$\V Q$} remain fixed.}
	\label{fig:Diagram}
	\vspace{-.5cm}
\end{figure}

% 3. why deep learning
%With the fast development of artificial intelligence in recent years, the Deep learning on image has drawn great success. While as the updating of sensors, more people start to focusing on the 3D world.

Since PCR is one of the most important functions in 3D perception, it also draws more attention to the Deep Learning (DL) community.
The DL research around PCR can be classified into two categories: (a) feature matching registration and (b) global feature registration.

%DONE: In the following paragraph, you should also cite your previous MDPI work
	% 3.1 local feature
Feature matching registrations rely on point detectors \cite{li2019usip,yew20183dfeat,lu2020rskdd} and descriptors \cite{yew20183dfeat,lu2020rskdd,bai2020d3feat,yuan2021self}, then point matching is applied to solve the transformation.
Very large rotations can be handled by matching-based registration~\cite{yang2020teaser}.
Anyhow, they are usually considered as a coarse registration to provide an initial guess for fine registration.
In hybrid models~\cite{choy2020deep,lu2021hregnet}, they assemble different registration modules to achieve a coarse-to-fine result.
%	
	% 3.2 global registration
In contrast to	feature matching-based registrations, global feature registrations are more direct.
For example, a network model has been designed as a black-box to take a pair of frames as input and output transformation result~\cite{choy2020deep}.	
%	
	% 3.3 global feature registration
More specific point cloud registrations using global features of each point cloud have also been explored~\cite{aoki2019pointnetlk,huang2020feature,li2021pointnetlk}.
Those methods use a global feature (vector) to represent each point cloud, which does not require correspondences for registration. 

We consider the global feature registration as highly interesting, because it eases the computation of point matching between a new arrived keyframe and batches of the previous $k$-frames.
In addition, the global features are kept for future uses.
%
% 4. shortage of deep learning
To compare with the conventional method, DL-based algorithms require a large amount of time for training.
When a new collection of data is obtained, those DL-based models cannot directly be used because the unseen data might not contain the same context as the trained dataset. Which results in an arbitrary guess.
%FIXED: Please state why
 
% 5. introduce our method

In this paper, we introduce a pseudo point-based method.
Following the diagram given in is shown in \cref{fig:Diagram}: 
Assume we have a function $\V \varphi$ that is transformation invariant with an arbitrary scan \textcolor{PineGreen}{$\V P$} and a pseudo point set \textcolor{Purple}{$\V S$} as input. Thus $\varphi$ feature is unchanged when both \textcolor{PineGreen}{$\V P$} and \textcolor{Purple}{$\V S$} apply the same transformation. Then a transformation between the point cloud \textcolor{PineGreen}{$\V P$} and \textcolor{Orange}{$\V Q$} is equivalent to transforming \textcolor{Dandelion}{$\V S'$} (in $\varphi(\textcolor{PineGreen}{\V P},\cdot)$) in the inverse direction.
Applying a similar registration framework to feature metric class registrations (DL methods), our model achieves similar or even better performance without any training.
In addition, feature metric class registrations are not compatible with real-world scenes~\cite{aoki2019pointnetlk,li2021pointnetlk}.
But our method is applicable in both indoor and outdoor settings.

%Another possible solution is to directly utilize Feature-metric framework on the latent feature of implicit function. But the constraint is the current recent research of distance field is with a large group of voxels because of the incapability of one feature to represent whole scene. Thus it is inefficient to generate a large groups of features during registration.

The contributions of this paper are:
\begin{itemize}
\item
  We propose a new direction of registration that registers a third pseudo point cloud while keeping the pairs of point clouds fixed.
  Thus the distance fields are also fixed. 
\item
  We propose the first implicit-to-implicit registration algorithm that does not require volumetric representation of the fields.
  I.e., we register two distance fields (or implicit-to-implicit) with sparse representations of the field.
\item
  We generalize the feature-metric registration framework to a non-DL setting.
	%\item introduces the Psudo-point Analytical Jacobian that circumvent the high space complexity problem of feature Jacobian in PointNetLK-Revisited~\cite{li2021pointnetlk}.
	%\item is compatible with synthetic, real point clouds.
\end{itemize}

%The advantage over FMR\cite{huang2020feature} is 1. only pass point cloud $P_S$, $P_\tau$ into encoder 3 times. 2. can work on large dataset.

%The advantage over \cite{zhu2021correspondence}, can have rotation and translation, also can work on large point cloud.

%\subsection{Overview}
%In the following, we discussed the related research: Feature Metric Registrations and Distance Field in \cref{sec:RelatedWork}. Then in \cref{sec:Method} our model is introduced. We evaluate our model in \cref{sec:Exp} and conclude this work in the end.

%Dimension Independent Global Feature for Point Cloud Registration 

%(Matching between two distance fields)
%(Class ourself into the feature-metric registration)

%Why do this work: 1. the feature-metrics global feature have to operate point cloud multiple times. 2. Not works on complex scene. 

%Storyline: 0. the introduce of feature-metrics registration and its advantage (O(NK)); 1. the feature-metrics global feature have to operate point cloud multiple times which is inefficient on large scale data; 2. dimension independent global feature for point cloud registration(O(KlogN)+O(N)); 3. our implementation can also works on complex scene.

\section{Related Work}
\label{sec:RelatedWork}

\subsection{Implicit-to-implicit registration}

A Signed Distance Field (SDF) is an implicit function $f:\mathbb{R}^3\rightarrow \mathbb{R}$ that maps a point $\V p\in \mathbb{R}^3$ to its closest surface location with signed distance~\cite{slavcheva2016sdf}.

Point-to-implicit registrations project point clouds to a SDF to circumvent the computation of correspondences.
It shows a more robust registration performance than ICP and is well-compatible with SDF-based 3D reconstruction algorithms.
But it suffers from the unreliability of sparse data and erroneous distance fields. 
SDF-2-SDF, as the first implicit-to-implicit registration algorithm, minimizes the energy between two dense volumetric SDFs, yielding a larger convergence radius and more accurate performance~\cite{slavcheva2016sdf}.
However, limited to the grid structure, SDF-2-SDF is restricted to be used in small settings.
To enable a large-scale scene reconstruction, SDF-TAR performs registration on a fixed number of limited-extent volumes with parallel computation \cite{slavcheva2016sdf2}.

But the above methods rely on an explicit volumetric representation of the implicit function.
It is in-efficient to recompute the field in each iteration.
To this end, we propose to register with two fixed distance fields, by introducing a third pseudo point cloud into the registration process.
%Recent years, the implicit function for distance field draw more attention in the area of reconstruction. 
%Neural unsigned distance field (NDF) ~\cite{chibane2020neural}.

\subsection{Feature Metric Registration}

Feature Metric Registrations (FMRs) are the group of algorithms for global feature registration with currently best performances. 
PointNetLK first introduced a registration framework with the feature distance~\cite{aoki2019pointnetlk}.
By iteratively updating the intermediate transformation, PointNetLK achieves high accuracy on synthetic datasets. 

Then FMR contributes on top of it and a semi-supervised algorithm was proposed in \cite{huang2020feature} to release the model from category labels.
In addition, this algorithm works well on indoor datasets.
However, those two models are using an approximate Jacobian, which means the step size should be carefully designed for avoiding too large or small steps along the gradient.

Recently, \cite{li2021pointnetlk} propose an analytical Jacobian for replacing the approximate Jacobian in PointNetLK.
The analytical Jacobian consists of a feature gradient and warp Jacobian.
They claim that the analytical Jacobian circumvents the numerical limitations.

%\begin{figure}[b!]
%	\centering
%	%\fbox{\rule{0pt}{2in} \rule{0.9\linewidth}{0pt}}
%	\includegraphics[width=1\linewidth]{im/LK.png}
%	\caption{}
%	\label{fig:LK}
%\end{figure}

In FMRs, iterating relies on the transformation of the whole point cloud $\mathbf P$, which is inefficient.
In this paper, we propose to directly update on the features. %In this way, the point cloud only requires constant pass for Jacobin $\mathbf J$ and the initial feature $\phi(\mathbf P_S)$, $\phi(\mathbf P_\tau)$.

\section{Background}

\subsection{Feature Metric Registrations}

Given two point clouds $\V P_S\in \mathbb{R}^{N_S\times 3}$, $\V P_\tau\in \mathbb{R}^{N_\tau\times 3}$, registration aims at finding the rigid transformation $\V G(\xi)$ such that $dist(\V G(\xi)\cdot \V P_S, \V P_\tau)$ is minimized. Here
$\V G(\xi)$ denotes the transformation with twist parameters $\xi$.

For feature metric registration, one minimizes the feature distance 
\begin{align}
  \arg\min_{\xi} \| \phi (\V G(\xi)\cdot \V P_S) - \phi(\V P_\tau) \| _2^2 .
\end{align}
where $\phi:\mathbb{R}^{N\times 3}\rightarrow \mathbb{R}^K$ denotes the encoding function with point number $N$ and global feature dimension $K$.

Then by rewriting the transformation on $\V P_\tau$ side, their warp increment $\Delta \xi$ is obtained by
\begin{align}
  \arg\min_{\Delta\xi} \| \phi (\V P_S) - \phi(\V G^{-1}(\xi^i \circ^{-1}\Delta\xi)\cdot \V P_\tau) \| _2^2 , 
	\label{eq:min_LK}
\end{align}
where $\circ^{-1}$ is an inverse composition.
By using the first-order Taylor expansion, it yields
\begin{align}
  \arg\min_{\xi} \| \phi (\V P_S) - \phi(\V G^{-1}(\xi^i)\cdot \V P_\tau) -\mathbf J\Delta \xi \| _2^2 ,
	\label{eq:min_latent_j}
\end{align}
where
\begin{align}
	\mathbf J=\frac{\partial  \phi(\V G^{-1}(\xi)\cdot \V  P_\tau)}{\partial \xi^T} \in \mathbb{R}^{K\times 6} .
\end{align}

%Then each twist parameter Jacobin $J_p\in\mathbb R^K$ is approximated with 
%\begin{equation}
%	\mathbf J_p\approx \frac{ \phi(G^{-1}(t_p)\cdot P_\tau)-\phi(P_\tau)}{t_p} \in \mathbb{R}^{K\times 6}
%\end{equation}
%where $t_p$ is the step size.
%There are two options to implement the Jacobian: the approximate Jacobian \cite{huang2020feature} and the analytical Jacobian \cite{li2021pointnetlk}, which are dicussed next. 
Recently an analytical Jacobian has been proposed in~\cite{li2021pointnetlk}. 
%FIXED: Check this sentance - English is not correct and I am not getting it
It circumvents the numerical unstable of approximate Jacobian~\cite{aoki2019pointnetlk} with step choices.
The analytical Jacobian consists of the Feature Jacobian and the Warp Jacobian:
\begin{align}
  \M J = \frac{\partial \phi(\V G^{-1}(\xi)\cdot \V P_\tau)}{\partial (\V G^{-1}(\xi)\cdot \V P_\tau)^{T}} \frac{\partial (\V G^{-1}(\V \xi)\cdot \V P_\tau)}{\partial \xi^{T}}.
\end{align}
Then by solving
\begin{align}
	\xi = \mathbf J^+[\phi(\V P_S)-\phi(\V P_\tau)]
	\label{eq:iter_s1}
\end{align}
we obtain the transformation and update as 	
\begin{align}
	\V P_S\leftarrow \V G(\xi)\cdot \V P_S . 
	\label{eq:iter_s2}
\end{align}
In the iteration, we solve \cref{eq:iter_s1} and \cref{eq:iter_s2} alternatively to make the final registration as
\begin{align}
	\V G = \Delta \V G_n \cdot \cdots \Delta \V G_0 ,
\end{align}
where $\Delta \V G = \exp(\sum_i \xi_i \V T_i)$.

\bigskip

\section{Methodology}
\label{sec:Method}

In each iteration, \cref{eq:iter_s2} always requires passing the original point cloud $\V P_S$ into the encoder function, which takes $\mathbb O(N \cdot C)$ where $N$ and $C$ are the maximum iterations and the number of points respectively.
It suffers from a similar problem as described in \cite{slavcheva2016sdf,slavcheva2016sdf2}, i.e., it requires to repeat operations on the original point clouds. 

We consider this transformation on point cloud to be inefficient and propose to solve it by using another small point set to ``communicate'' with $\V P_S$ and $\V P_\tau$ instead of letting $\V P_S$ and $\V P_\tau$ ``contact'' directly.
We use a random sampled $L$-points point cloud $\V P_a\in \mathbb{R}^{L\times 3}$ where $L\ll N$. %Here we pre-assume that $\V P_S$ and $\V P_\tau$ are in a $2s$-cubic.

Then we use the encoding function $\varphi(\cdot,\V P_a)$ with $\varphi:\mathbb{R}^{N\times 3}\times \mathbb{R}^{L\times 3}\rightarrow \mathbb{R}^{K}$ to extract features from the point cloud with $\V P_a$ involved. So $\varphi(\V P_S,\V P_a)$ between $\V P_S$ and $\V P_a$ is the $\V P_S$ feature.
%FIXED: We use the feature to represent the feature? Sounds awkward.
$\varphi(\V P_\tau,\V P_a)$ is the feature extracted from $\V P_\tau$. 

Here we assume $\varphi(\V P_S,\V P_a)$ is \textbf{transformation invariant} to $\V P_S$ and $\V P_a$ as in \cref{fig:Diagram}, such that the transformation on $\V P_S$ is equivalent to the inverse transformation on $\V P_a$, i.e.,
\begin{align}
	\varphi (\V G(\xi)\cdot \V P_S,\V P_a) = \varphi (\V P_S,\V G^{-1}(\xi)\cdot \V P_a).
	\label{eq:transformation_invariant}
\end{align} 

\subsection{Distance Field as a Transformation Invariant Function}

\begin{figure}[b]
	\vspace{-.2cm}
	\centering
	\begin{subfigure}{.25\textwidth}
		\centering
		\includegraphics[width=.8\linewidth]{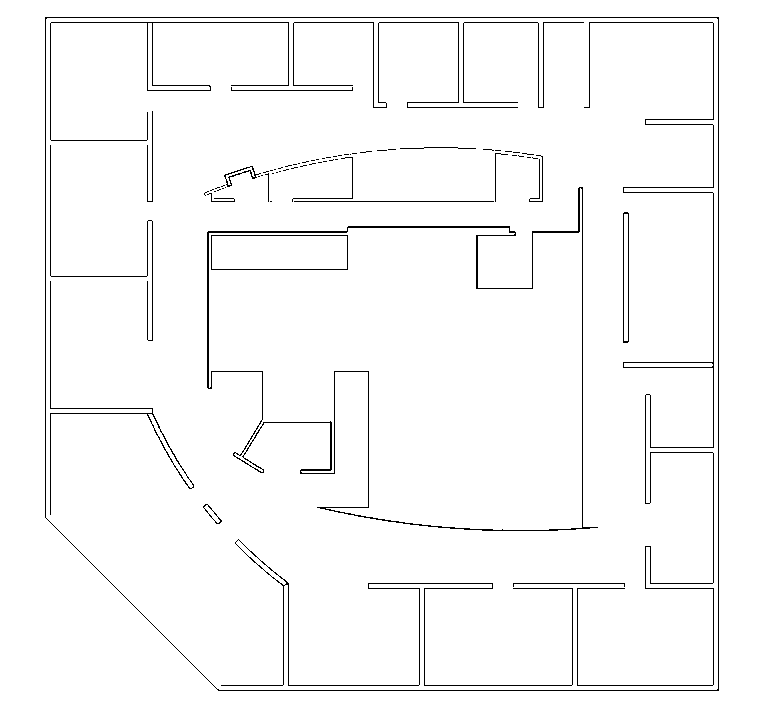}
		\caption{Intel 2D map.}
	\end{subfigure}%
	\begin{subfigure}{.25\textwidth}
		\centering
		\includegraphics[width=.8\linewidth]{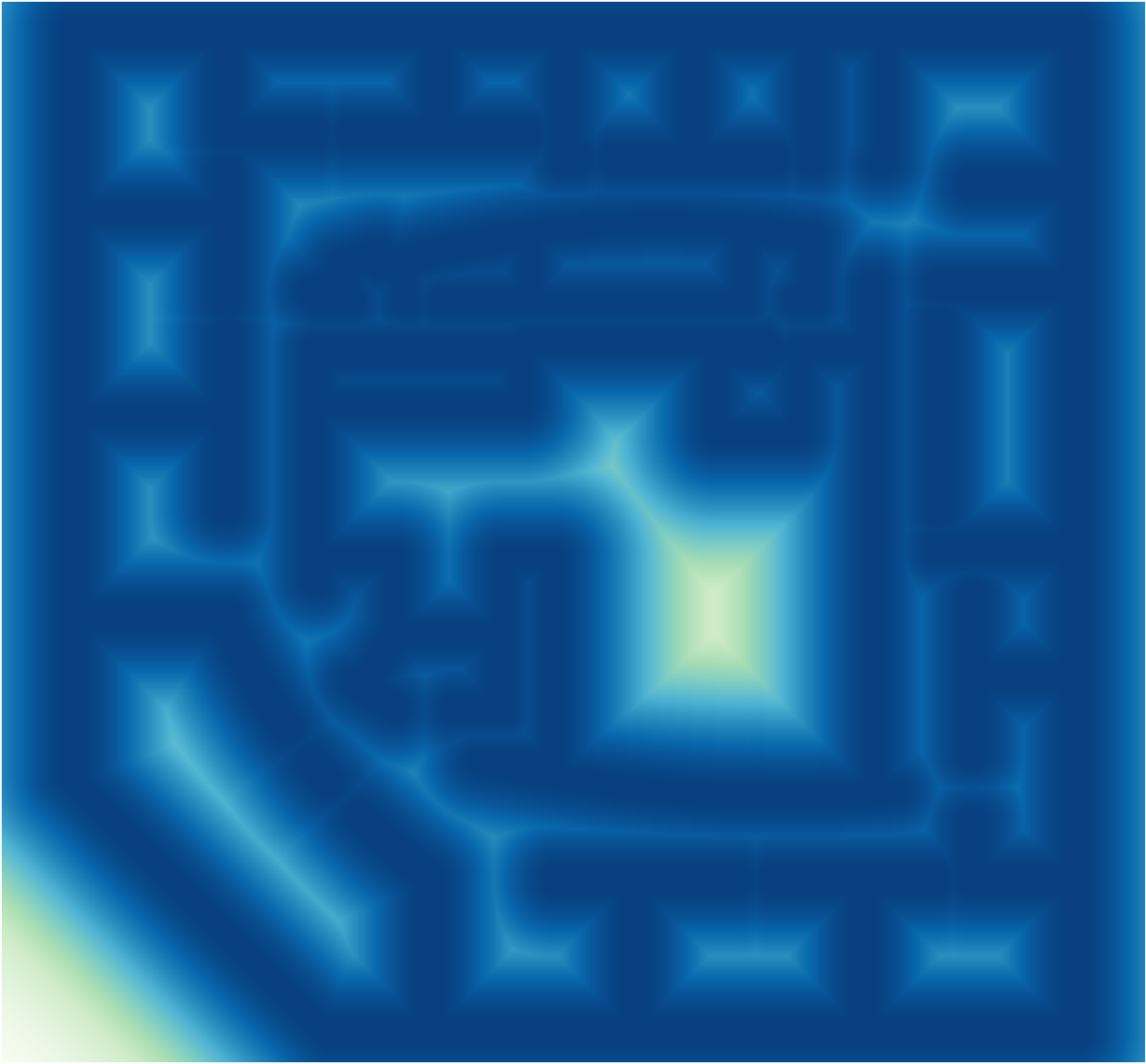}
		\caption{Distance field map.}
	\end{subfigure}%
\\
	\begin{subfigure}{.25\textwidth}
	\centering
	\includegraphics[height=.7\linewidth]{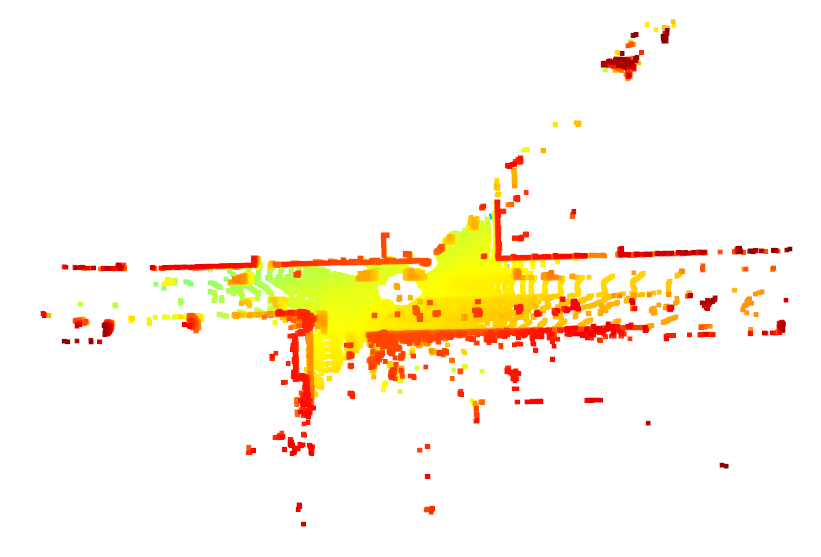}
	\caption{KITTI scan.}
\end{subfigure}%
\begin{subfigure}{.25\textwidth}
	\centering
	\includegraphics[height=.7\linewidth]{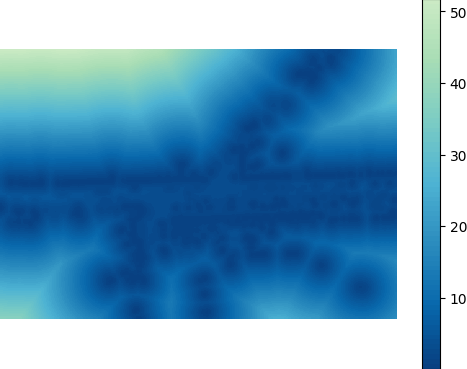}
	\caption{Distance field 2D sketch.}
\end{subfigure}%
	
	\caption{Distance field demonstrations. The above is for 2D Intel map generated with \cite{yuan2019incrementally}. Below is on selected KITTI-odometry frame. }
%FIXED: Can you put here also an example from the KITTI data set. I.e., compute a sample distance dield map from one of the point clouds.
	\label{fig:distance}
\end{figure}

The formulation in \cref{eq:transformation_invariant} holds only when each dimension value of $\varphi(\V P_S,\V G^{-1}\V P_a)$ will not change after rotation and translation on the inputs of $\varphi$.
Thus, in the view of the feature metric, where $\varphi(\V P_\tau,\V P_a)$ and the transformed $\varphi(\V P_S,\V G^{-1}\V P_a)$ are compared,
we cannot separately use the feature of $\V P_S / \V P_\tau$ or $\V P_a$. Because we neither transform $\V P_S / \V P_\tau$ in our iteration design, nor $\V P_a$ gives anything itself. Therefore, we consider $\V P_a$ as a medium to help represent the geometry of $\V P_S / \V P_\tau$.
%we cannot use point set feature of $\V P_a$ in $\varphi$ since the dimensions of deep feature are not independent to each other. 
%
We observe that the point order of $\V P_a$ is actually fixed for $\varphi(\V P_S,\V G^{-1}\V P_a)$ and $\varphi(\V P_\tau,\V P_a)$ in one registration process. Which provides independence in each dimension and makes the algorithm design more flexible (in \cref{sec::IRLS} and \cref{sec:trunc}).
In addition, the function for each point should provide continuous values for derivation computation.

We propose to introduce the distance field and use the score of $\V p_i\in \V P_a$ in the field $\V D_{\V P_S}$ as the $i$-th dimension of $\varphi(\V P_S,\V P_a)\in\mathbb R^{L} $.
The concept of the distance field is then utilized for the transformation invariant function with the input of scan and pseudo points.
Since an unsigned distance field is able to work on an arbitrary point cloud, that is not limited to a closed shape, we implement our algorithm under this more generalized setting.
The distance field is a function that computes the smallest distance between query point $\V x$ and the point cloud $\V P$
\begin{align}
	\V D_{\V P}(\V x)=\min_{\V p\in\V  P}(||\V  p-\V x  ||_2^2) .
\end{align}

The demonstration of distance field is in \cref{fig:distance}.

\subsection{Pseudo-point Feature Metric Registration}
\label{sec:method:1}

%\begin{figure*}[t]
%	\centering
%	%\fbox{\rule{0pt}{2in} \rule{0.9\linewidth}{0pt}}
%	\includegraphics[width=0.85\linewidth]{im/moon3.png}
%	\caption{Algorithm pipeline of proposed method. The \textcolor{CarnationPink}{pink} arrows denote computations that are carried out once. The \textcolor{Goldenrod}{yellow} arrows denote computations in an iterative fashion.}
%	\label{fig:LK}
%\end{figure*}

\begin{figure*}[t]
	\centering
		\psfragfig[width=.8\textwidth,height=.4\textwidth]{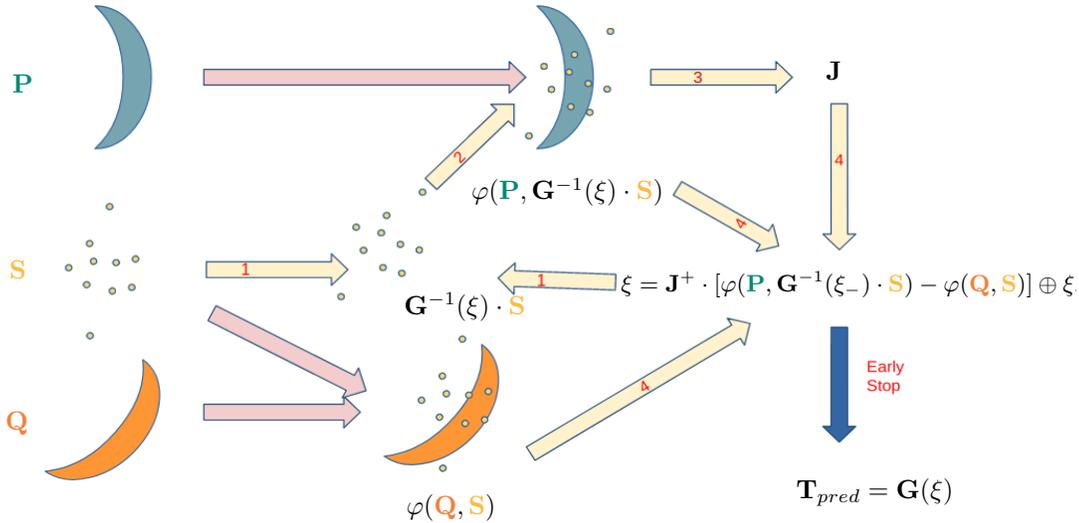}{
	\psfrag{J}{$\V J$}
	\psfrag{P}{$\textcolor{PineGreen}{\V P}$}
	\psfrag{S}{$\textcolor{Dandelion}{\V S}$}
	\psfrag{Q}{$\textcolor{Orange}{\V Q}$}
	\psfrag{TP}{$\V T_{pred}=\V G(\xi)$}
	\psfrag{GS}{$\V G^{-1}(\xi)\cdot \textcolor{Dandelion}{\V S}$}
	\psfrag{F(QS)}{$\varphi(\textcolor{Orange}{\V Q},\textcolor{Dandelion}{\V S})$}
	\psfrag{F(P,GS)}{$\varphi(\textcolor{PineGreen}{\V P},\V G^{-1}(\xi)\cdot \textcolor{Dandelion}{\V S})$}
	\psfrag{F(PGS)-F(QS)SSS}{\small$\xi = \V J^{+}\cdot [\varphi(\textcolor{PineGreen}{\V P},\V G^{-1}(\xi_{-})\cdot \textcolor{Dandelion}{\V S})-\varphi(\textcolor{Orange}{\V Q},\textcolor{Dandelion}{\V S})]\oplus \xi_{-}$}
	 }
%FIXED: Explain the numbers 1 ... 4 in the caption. And maybe also in the text.          
	\caption{Algorithm pipeline of the proposed method. The \textcolor{CarnationPink}{pink} arrows denote computations that are carried out once. The \textcolor{Goldenrod}{yellow} arrows denote computations in an iterative fashion. The \textcolor{red}{1}, \textcolor{red}{2}, \textcolor{red}{3}, \textcolor{red}{4} are sequential steps in an iteration. We show algorithm of this pipeline in supplementary material$^2$.}
\label{fig:LK}
	\vspace{-.5cm}
\end{figure*}
Given above function $\varphi$, shown similarly in \cref{fig:Diagram}, the transformation between $\V P_S$ and $\V P_\tau$ becomes the \textbf{inverse-alignment} of $\V P_a$ to make $\varphi(\V P_S,\V G^{-1}\cdot \V P_a)$ and $\varphi(\V P_\tau, \V P_a)$ equal. Which is 
\begin{align}
	\arg\min_{\xi} \| \varphi (\V P_S,\V G^{-1}(\xi)\cdot \V P_a) - \varphi(\V P_\tau,\V P_a) \| _2^2
	\label{eq:PFMR}
\end{align}

%A pseudo-point feature is a sparse representation of the distance field and will be formally introduced in next subsection.
By computing the Jacobian one is not able to describe the gradient on the whole field.
Different from \cref{eq:min_LK,eq:min_latent_j} that rewrite the transformation onto the target $\V P_\tau$ side.
Instead, we keep updating the feature of $\V P_S$ and update Jacobian in each iteration.
\cref{eq:PFMR} is rewritten by adding an incremental $\Delta \xi$ as
\begin{align}
	\arg\min_{\Delta\xi} \| \varphi (\V P_S,\V G^{-1}(\xi^i \circ^{-1}\Delta\xi)\cdot \V P_a) - \varphi(\V P_\tau, \V P_a) \| _2^2 .
	\label{eq:min_LK_r}
\end{align} 
Using the first-order Taylor expansion, \cref{eq:min_LK_r} becomes
\begin{align}
	\arg\min_{\xi} \| \varphi (\V P_\tau,\V P_a) - \varphi(\V P_S,\V G^{-1}(\xi^i)\cdot \V P_a) -\V J\Delta \xi \| _2^2
	\label{eq:min_latent_j_r}
\end{align}
%with Jacobian
%\begin{equation}
%	\mathbf J=\frac{\partial  \varphi(P_\tau, G(\xi)\cdot P_r)}{\partial \xi^T} \in \mathbb{R}^{K\times 6}.
%\end{equation}

In our algorithm, the Jacobian is computed in each iteration. %Approximate Jacobian is intractable as it requires seven computations of $\varphi$.
%So in this paper, the analytical Jacobian is formulated.
%
The analytical Jacobian is formulated. It contains the feature Jacobian over the pseudo-point set and multiplies the warp Jacobian of pseudo-point set over the transformation parameters:
\begin{align}	
\V  J = \frac{\partial\varphi(\V P_S,\V G(\xi^i)^{-1}\cdot \V P_a)}{\partial (\V G(\xi^i)^{-1}\cdot \V P_a)^{T}} \frac{\partial ( \V G(\xi^i)^{-1}\cdot \V P_a)} {\partial \xi^T}.
\end{align}
%To note that, instead of computing the Jacobian related to original point cloud as in \cite{li2021pointnetlk}, we compute the intermediate Jacobian for the pseudo-point set $\V P_a$. 
%
The warp Jacobian $\frac{\partial ( \V G(\xi^i)^{-1}\cdot \V P_a)} {\partial \xi^T}\in \mathbb{R}^{L\times 3\times 6}$ follows to the corresponding formulation on the pseudo-point set $\V P_a$ as PointNetLK-Revisited~\cite{li2021pointnetlk} uses it for the original point set.
The feature Jacobian $\V J_{feat}= \frac{\partial\varphi(\V P_S,\V G(\xi^i)^{-1}\cdot \V P_a)}{\partial (\V G(\xi^i)^{-1}\cdot \V P_a)^T} \in \mathbb{R}^{L\times 3\times L}$ is the gradient with closest points of pseudo-points in our implementation:
\begin{align}
	\begin{split}
	\V J_{feat,(i,\cdot,i)} &= 2(\V x_i-\V p) \in \mathbb{R}^{3},\\
	&\text{with}\   \V x_i\in \V P_a,\  \V p=\argmin_{\V p\in \V P_S}(|| \V p-\V x_i  ||_2^2) .
	\end{split}
	\label{eq:Jacobain}
\end{align}
And the remaining non-filled entries of $\V J_{feat}$ are set to zero. Our pseudo-set analytical Jacobian is far more efficient than the analytical Jacobian of \cite{li2021pointnetlk} and is discussed in supplementary material\footnote{See github page$^{1}$}.%\cref{sec:advantage}.

Then by following \cref{eq:iter_s1}, $\xi$ is solved with
\begin{align}
	\label{eq:solve_xi}
	\xi = \mathbf \V J^+[\varphi(\V P_\tau,\V P_a)-\varphi(\V P_S,\V G^{-1}(\xi_{-})\cdot \V P_a)] \oplus \xi_{-}
\end{align}
where $\oplus$ is denoting the update between two twist vectors, $\xi_{-}$ is denoting the twist parameters from last iteration.
%
%In each iteration, the updating becomes
%\begin{equation}
%	\varphi(P_S,P_a)\leftarrow \varphi(P_S,G^{-1}(\xi)\cdot P_a)
%	\label{eq:iter_s2_r}
%\end{equation}
%And the $\varphi(P_\tau,P_a)$ is fixed during iteration.
%
The whole diagram of the registration is shown in \cref{fig:LK}.
Before the first iteration, $\varphi(\V P_S,\cdot)$ and $\varphi(\V P_\tau,\V P_a)$ are prepared. Then in each iteration $i$, we transform $\V P_a$ as $G^{-1}(\xi_i)\V P_a$ to compute the feature and Jacobian for $\varphi(\V P_S,\V G^{-1}(\xi_{-})\cdot \V P_a)$. Then the incremental transformation is solved with the linear system $\V J\xi=\V r$, where $\V r$ is the residual of features. 

\subsection{Robust regression with Weights}
\label{sec::IRLS}
Different from deep feature metric registration that dimensions are correlated to each other, in our method each dimension of a feature is actually corresponding to a single pseudo-point.
This means each dimension is independent to the others.
Thus, we are able to add weight to each dimension of the function $\mathbf J\xi=\mathbf r$ to adjust the importance of each dimension. 
Therefore, we implement the Iterative Reweighted Least Squares (IRLS) to solve for the $\xi$. With the weighted formulation, we actually solve the sum of $L_1$ version of \cref{eq:min_LK_r}. For more detail on IRLS, please follow \footnote{\url{https://en.wikipedia.org/wiki/Iteratively_reweighted_least_squares\#cite_ref-Burrus_1-0}} and \cite{gentle2007matrix}.
%FIXE: Better a real reference?

\begin{table*}[t!]
	\centering
	\caption{Accuracy and generalizability. Results on 20 object categories, i.e., unseen categories of the baseline learning methods, from ModelNet40 and ShapeNetCore. 
		$\downarrow$ means smaller values are better. }
	\begin{threeparttable}
		%\begin{adjustbox}{width=2\columnwidth}
			\centering
			\begin{tabular}{lcccccccc}\toprule
				& \multicolumn{4}{c}{ModelNet40} & \multicolumn{4}{c}{ShapeNetCore} \\ \cmidrule(lr){2-5} \cmidrule(lr){6-9}
				& \multicolumn{2}{c}{Rot. Err. (deg.)~$\downarrow$} & \multicolumn{2}{c}{Trans. Err.~$\downarrow$} & \multicolumn{2}{c}{Rot. Err. (deg.)~$\downarrow$} & \multicolumn{2}{c}{Trans. Err.~$\downarrow$} \\ 
				\cmidrule(lr){2-3} \cmidrule(lr){4-5} \cmidrule(lr){6-7} \cmidrule(lr){8-9}
				Algorithm & RMSE & Median & RMSE & Median & RMSE & Median & RMSE & Median \\ \midrule
				ICP$^*$~\cite{besl1992method} & 39.33 & 5.036 & 0.474 & 0.058 & 40.71 & 5.825 & 0.478 & 0.073 \\
				DCP$^*$~\cite{wang2019deep} & 5.500 & 1.202 & 0.022 & 0.004 & 8.587 & 0.930 & 0.021 & 0.003 \\
				DeepGMR$^*$~\cite{yuan2020deepgmr} & 6.059 & 0.070 & \textbf{0.014} & 8.42e-5 & 6.043 & 0.013 & \textbf{0.005} & 9.33e-6 \\
				PointNetLK$^*$~\cite{aoki2019pointnetlk} & 8.183 & 3.63e-6 & 0.074 & 5.96e-8 & 12.941 & 4.33e-6 & 0.115 & 5.96e-8 \\
				PointNetLK-Revisited$^*$~\cite{li2021pointnetlk}& \textbf{3.350} & {2.17e-6} & 0.031 & \textbf{4.47e-8} & \textbf{3.983} & \textbf{2.06e-6} & 0.049 & \textbf{2.98e-8} \\ 
				\midrule
				Ours & 5.168 & \textbf{1.83e-6} & 0.055 & \textbf{4.47e-8} & \textbf{4.042} &\textbf{2.95e-6} & 0.047 & \textbf{5.96e-8} \\  
				
				\bottomrule
			\end{tabular}
		%\end{adjustbox}
		\vspace{0.01ex}
		\begin{tablenotes}
			\item[$^*$] Values taken from \cite{li2021pointnetlk}.
		\end{tablenotes}
	\end{threeparttable}
	\label{tb:accuracy}
	\vspace{-.5cm}
\end{table*}

\subsection{Truncation Strategy}
\label{sec:trunc}
In the implementation, truncation is used to remove the outlier points as in each iteration, certain dimensions of $\varphi$ are freely removed together with its corresponding Jacobian.
For example, if the $j$-th pseudo-point is removed, $K^{+} = K-1$ for the feature $\varphi$ of both sample and template point cloud, then the $j$-th row of the Jacobian is also ignored in \cref{eq:iter_s1}. It is also feasible as it is considered as part of $\varphi$ function to block certain dimensions in the iteration.
The specific implementation of the truncation is described in \cref{sec::setting}.

\subsection{Computational cost}
The computation cost of our method includes the implementation of the distance function. With the recent neural network based distance field~\cite{park2019deepsdf,chibane2020neural}, it costs $\mathbb{O}(N)+\mathbb{O}(K L)$ to get the point cloud global latent and to iterate $K$ times for the pseudo-point set. 
In this paper, we focus on the algorithm itself, and compute the distance value for each pseudo-points by using $k$-d tree search.
Excluding the $\mathbb{O}(N \log N)$ tree-building time, the registration time complexity of our model is $\mathbb{O}(K L \log N)$.

In comparison, excluding the $\mathbb{O}(N \log N)$ tree building time, ICP takes $\mathbb{O}(K N \log N)$ for iterating $K$ times.
Feature metric registrations require $\mathbb{O}(K N)$.
More detail about space complexity is given in supplementary material$^2$.%~\cref{sec:advantage}.

\section{Experiments, Results, and Discussion}
\label{sec:Exp}

Next, we focus on the registration performance to evaluate our model on synthetic datasets, an indoor dataset, and an outdoor dataset.
The deep learning based models are known to exceed the conventional method on performance.
However, our results show that, even without any training of the model, as a feature-metric method, our model achieves competitive results to the state-of-the-art learned models. 

\subsection{Datasets}
\begin{figure}[b!]
	\vspace{-.5cm}
	\centering
	\includegraphics[width=.2\linewidth]{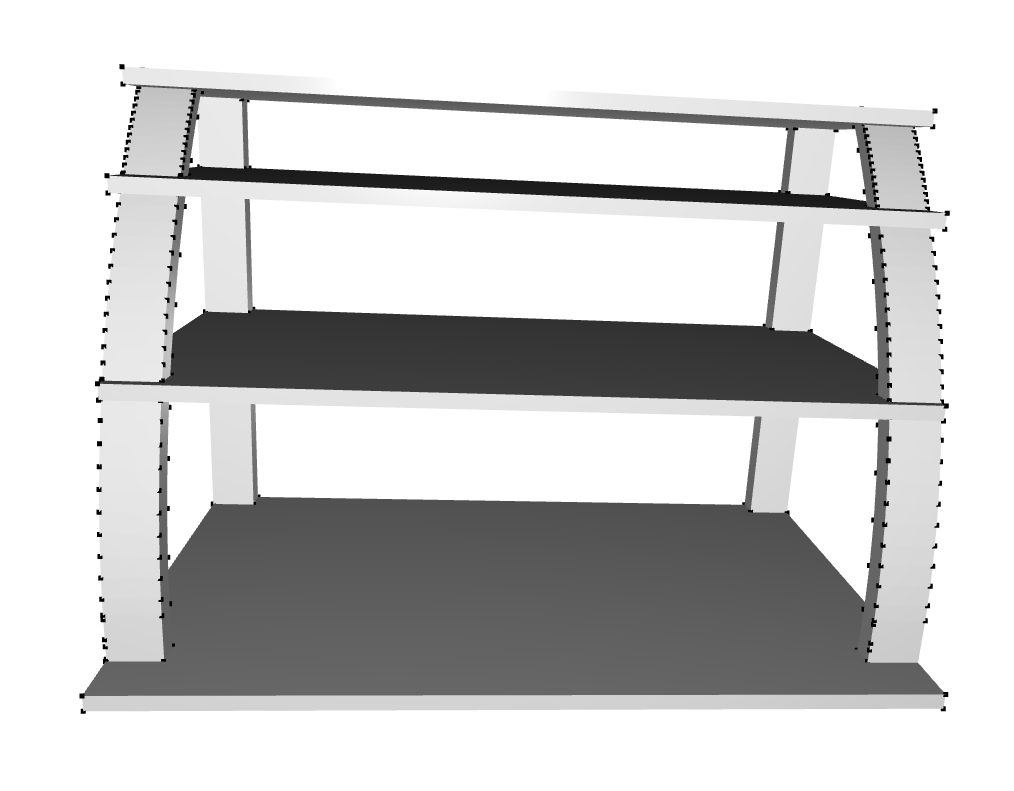}
	\includegraphics[width=.2\linewidth]{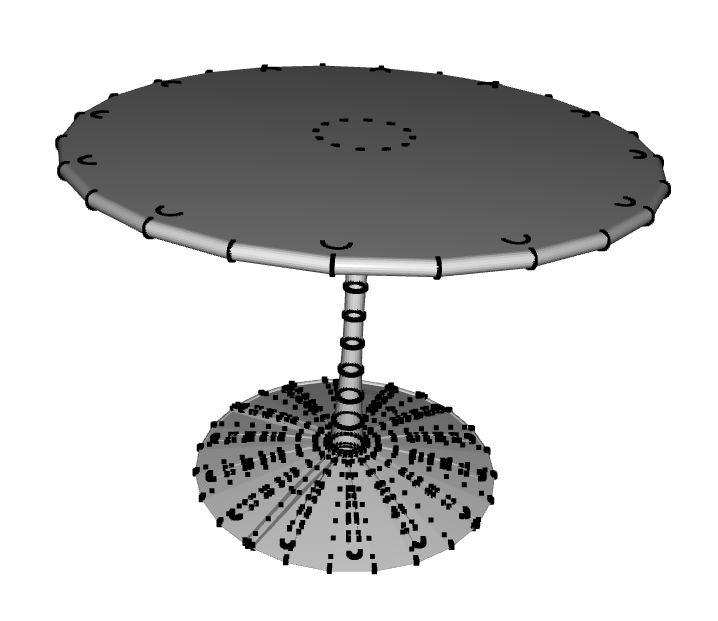}
	\includegraphics[width=.2\linewidth]{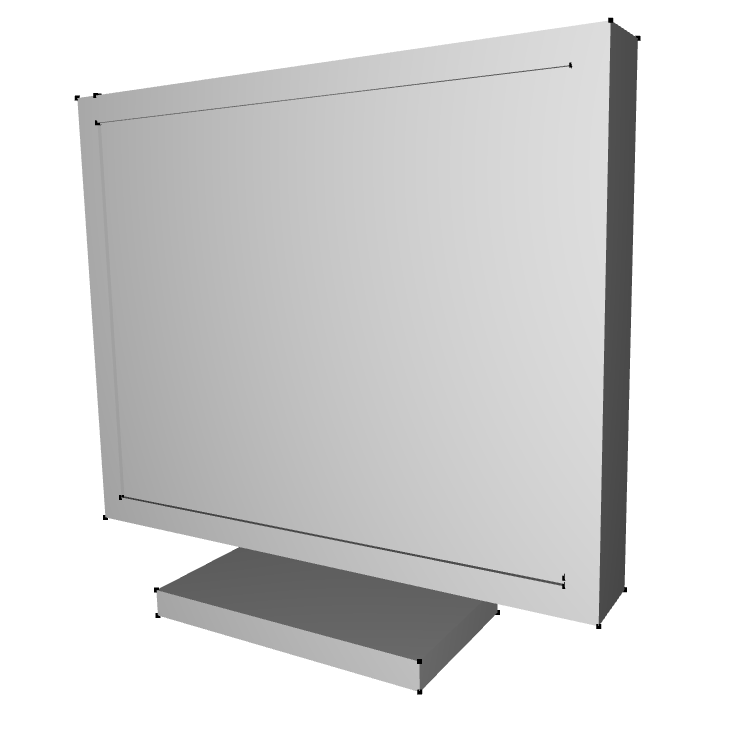}
	\includegraphics[width=.2\linewidth]{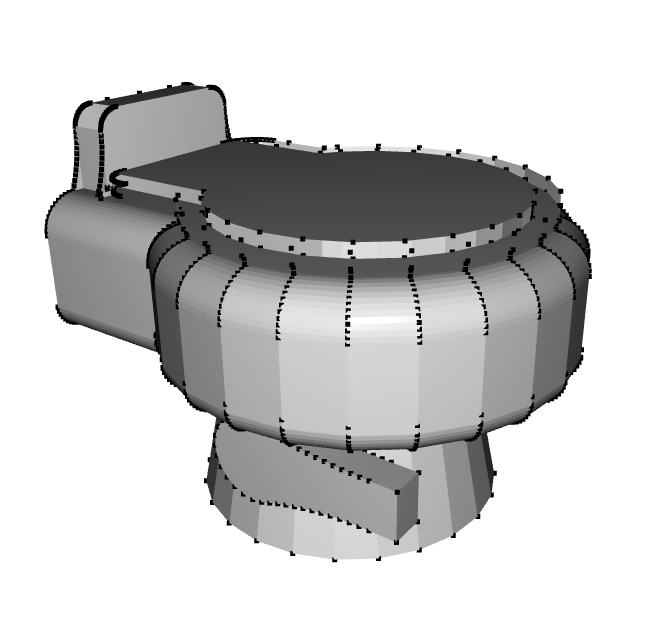}
	\\
	\includegraphics[width=.2\linewidth]{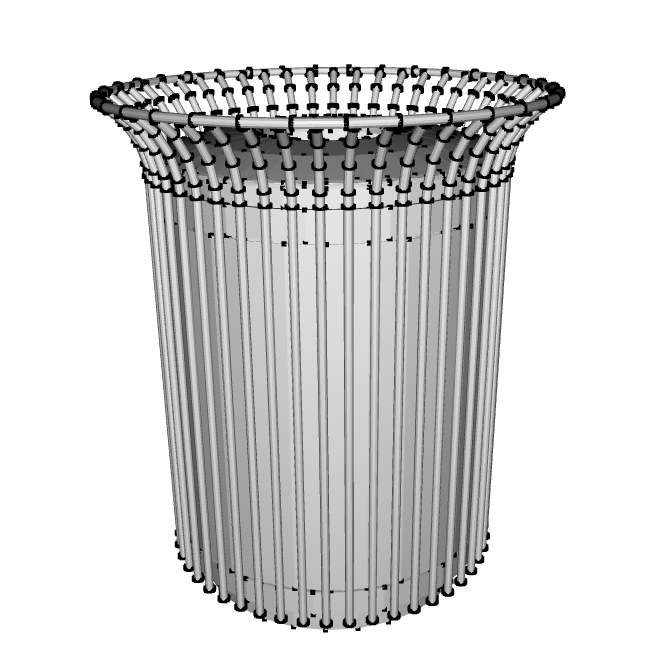}
	\includegraphics[width=.2\linewidth]{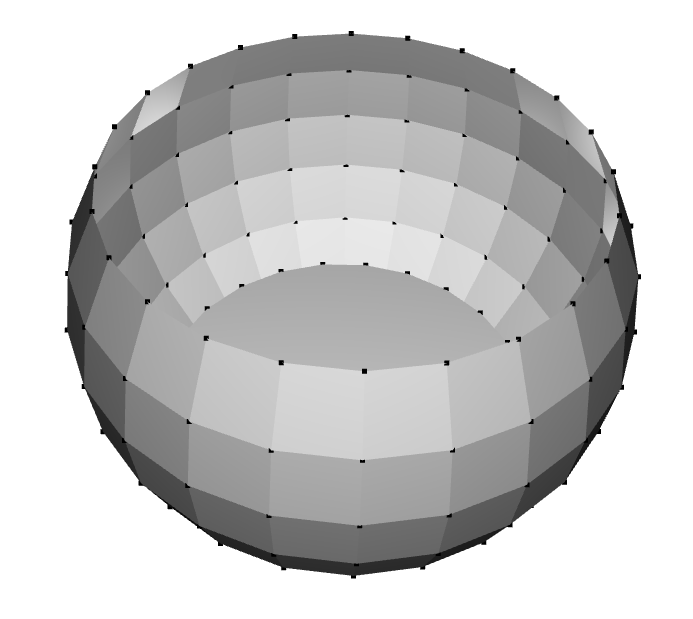}
	\includegraphics[width=.2\linewidth]{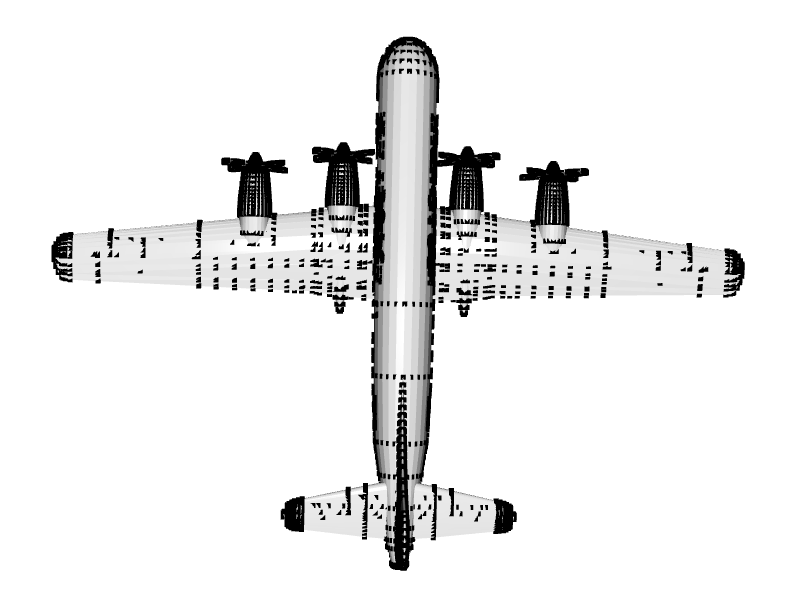}
	\includegraphics[width=.2\linewidth]{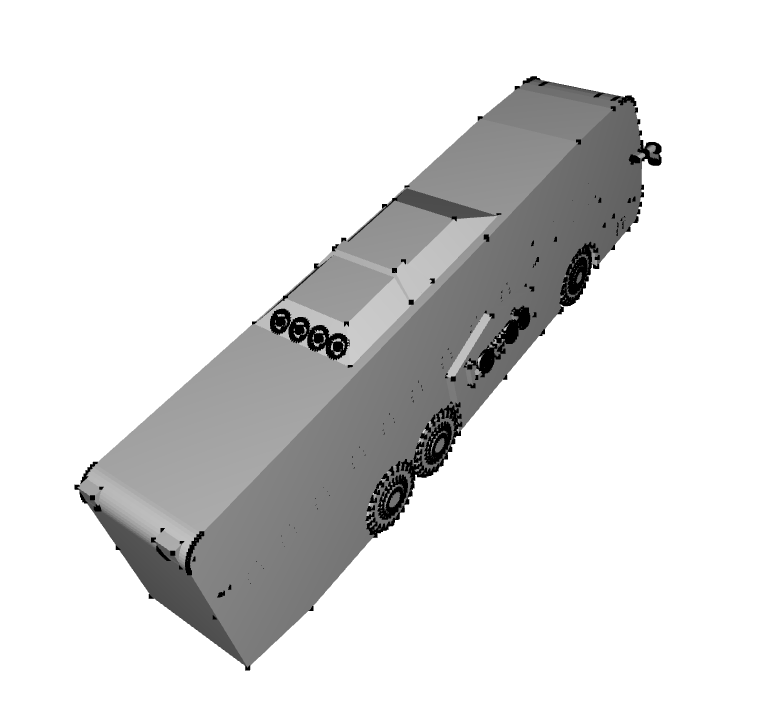}
	\\
	\includegraphics[width=.4\linewidth]{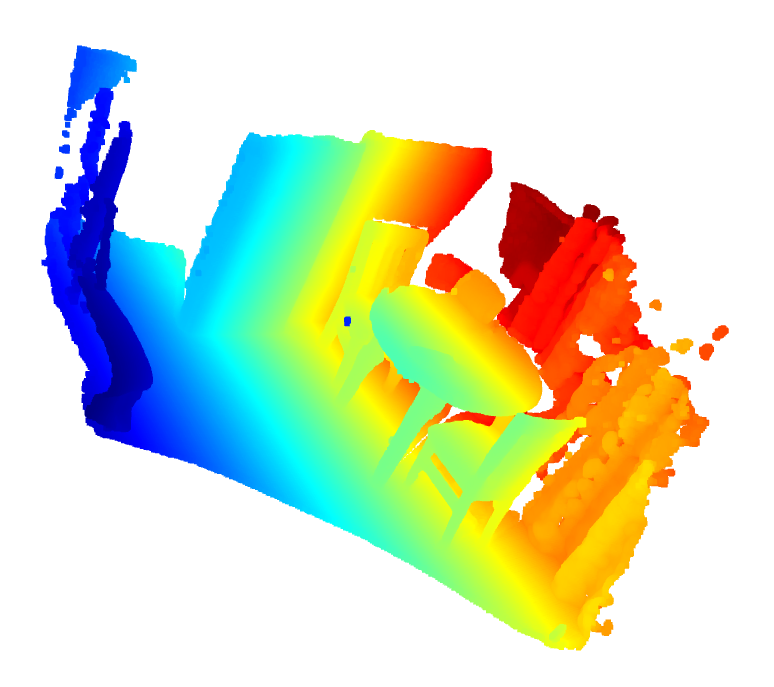}
	\includegraphics[width=.3\linewidth]{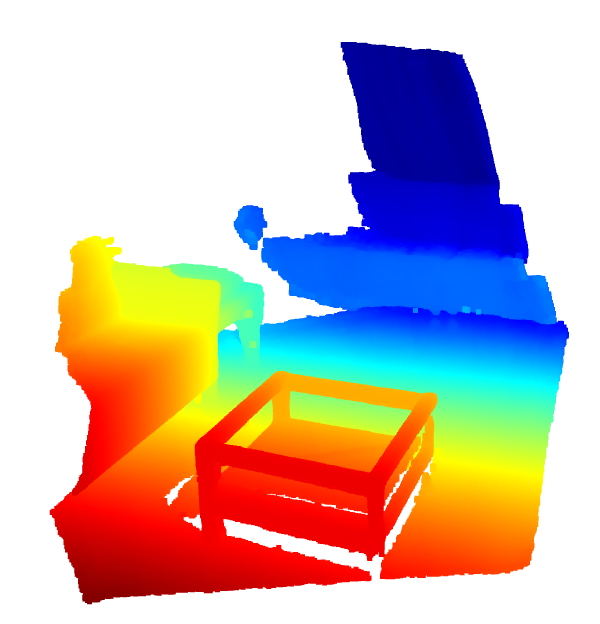}
	\\ 
	\includegraphics[width=.8\linewidth]{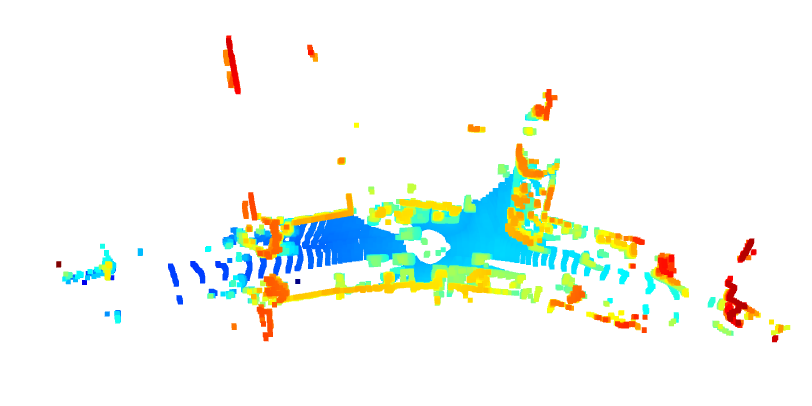}
	%\\ \vspace{-.5cm}
	%\includegraphics[width=.5\linewidth]{im/pcd/kitti/4.png}
	\caption{Data examples. The four rows are selected examples from ModelNet, ShapeNet, 3DMatch, KITTI-Odometry datasets respectively. In the first two rows, aside from the black point cloud we also plot its object surface as in~\cite{li2021pointnetlk} for better visualization.}    
	\label{fig:data}
	\vspace{-.5cm}
\end{figure}

\subsubsection{ModelNet40~\cite{wu20153d}} ModelNet40 is a synthetic 3D object dataset contains 40 categories of daily objects.
For a fair comparison, we follow PointNetLK-Revisited to select 20 categories from ModelNet40 as a testing set, the rest is for the training of other learning models.

\subsubsection{ShapeNetCoreV2~\cite{chang2015shapenet}} ShapeNetCore is a subset of ShapeNet that contains 55 common object categories.
Again, for a fair comparison, 12 categories are selected from ShapeNetCoreV2 separately for testing and the rest for the training of other learned models. 

\subsubsection{3DMatch~\cite{zeng20173dmatch}} 3DMatch collects RGB-D scans from several datasets.
It contains complex indoor scenes, such as kitchen, office, etc.
We follow PointNetLK-Revisited to choose 8 categories of scenes for testing.
In addition, only those testing pairs with at least 70\% overlap are kept. 

\subsubsection{KITTI-odometry~\cite{geiger2012we}} KITTI-odometry is a large-scale outdoor LiDAR point cloud dataset around cities. It consists of 22 sequences. The first 11 sequences are provided with ground truth trajectory. This test on outdoor data is used to compare with outdoor registration models to further explore the capacity of our algorithm.
%\subsubsection{NuScenes \cite{caesar2020nuscenes}} is the other widely used large scale outdoor dataset. It contains 1000 scenes, from which, 850 are officially splited for training and the rest 150 for testing.

%For the large scale outdoor datasets, we follow HregNet~\cite{lu2021hregnet}  to prepare data. For KITTI-odometry, we select 08 to 10 sequences for test. Pairs are generated with a 10 frames gap. In addition, it use ICP to refine the noise transformation. For the NuScene datasets, the 150 test split are involved for our test.

Selected examples of used data are shown in \cref{fig:data}.

\subsection{Baselines}
%FIXED: fix citation
We use as baseline ICP~\cite{besl1992method}, DCP~\cite{wang2019deep}, DeepGMR~\cite{yuan2020deepgmr}, PointNetLK~\cite{aoki2019pointnetlk}, and the most recent state-of-the-art PointNetLK-Revisited~\cite{li2021pointnetlk} for synthetic registration and indoor registration.
We also test it on an outdoor dataset which is not applicable to deep feature-based registration~\cite{li2021pointnetlk}. Thus we use the benchmark from outdoor registration paper~\cite{horn2020deepclr} with the baseline ICP~\cite{besl1992method}, G-ICP~\cite{segal2009generalized}, 3DFeat-Net~\cite{yew20183dfeat}, DeepVCP~\cite{lu2019deepvcp}, and DeepCLR~\cite{horn2020deepclr}.
\subsection{Metrics}
\vspace{-.1cm}
The difference between prediction and ground truth for rotation (in degree) and translation (in meter) is evaluated as relative rotation error
(RRE$=\arccos(Tr(R^T R-1)/2)$) and relative translation error (RTE).
To represent the variance of registration and the error distribution we use the root mean square error (RMSE) and median error (MAE) of RRE, RTE as metrics.
%For outdoor registration, which is challenge and more in the class of global registration, we compare with ICP, FGR, RANSAC, DCP, IDAM, FMR, DGR and the most recent outdoor global registration SOTA HregNet [CVPR2021].

\subsection{Setting}
\label{sec::setting}
The rigid transformation used for testing consists of a rotation randomly drawn in [\ang{0}, \ang{45}] and a translation in [\num0, \num0.8] for synthetic and indoor test~\cite{li2021pointnetlk}. The outdoor test follows~\cite{horn2020deepclr}.
As our model does not require training, we directly feed all of the points as input. 

The maximum number of iterations is set to 10 for iterative methods on synthetic data and to 20 on real data. The number of pseudo points $L$ is $1000$ in all experiments.

For pseudo set generation, we implement a random pseudo set and a neighborhood pseudo set generation for following experiments.
For truncation, we filtered out pseudo-points if (1) too many pseudo-points redundantly neighbors to the same surface point and (2) pseudo-points are not on the surface normal direction, as those points are more likely to be far away from the surface or in a no-surface region.
%Then the pseudo-point with non-unique neighbor index will not participate in the registration in the current iteration.
%
%\subsubsection{Pseudo-points not in the Normal Direction}
%In this strategy, each pseudo-point knows its neighbor.
%Then by comparing the direction vector $\overrightharp{d}$ and the neighbour normal $\overrightharp{n}$ with the threshold $\epsilon_{orient}$ we get
%$Mask[p] = | \overrightharp{d}^T \cdot \overrightharp{n} | > \epsilon_{orient}$ to filter out points in current iteration.

All experiments are implemented on a NUC-computer (CPU-i7-10710U, 32GB memory).
\subsection{Synthetic Data Test}
\label{sec:synthetic_test}
In the synthetic test, pseudo set are uniformly drawn from $[-1,1]^3$ cube. No truncation has been used.

%\subsubsection{Accuracy and generalizability}
The evaluation performance is shown in \cref{tb:accuracy}.
Our non-learning method (IFR) obtained a competitive performance to the new state-of-the-art PointNetLK-Revisited~\cite{li2021pointnetlk}.
The feature-metric class methods (PointNetLK, PointNetLK-Revisited, ours) achieves a median rotation error at \num{1e-6} level and a translation error at \num{1e-8} level, which is much more precise than the rest.

\begin{figure}[]
	\centering
	\begin{subfigure}[t]{.5\linewidth}
		\centering
		\includegraphics[width=.98\linewidth]{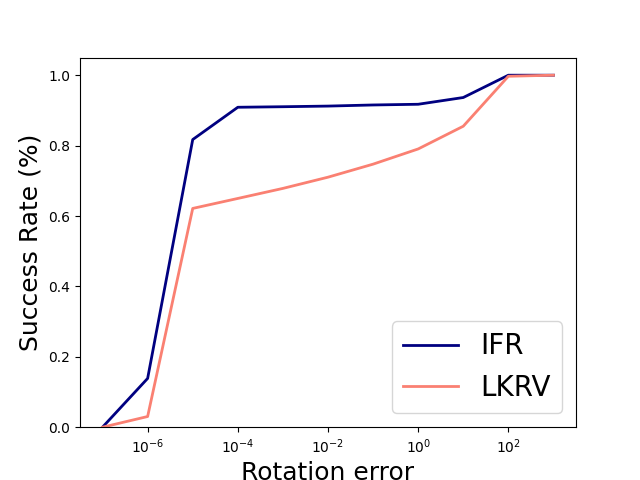}
		\caption{Rotation threshold. }
	\end{subfigure}~
	\begin{subfigure}[t]{.5\linewidth}
		\centering
		\includegraphics[width=.98\linewidth]{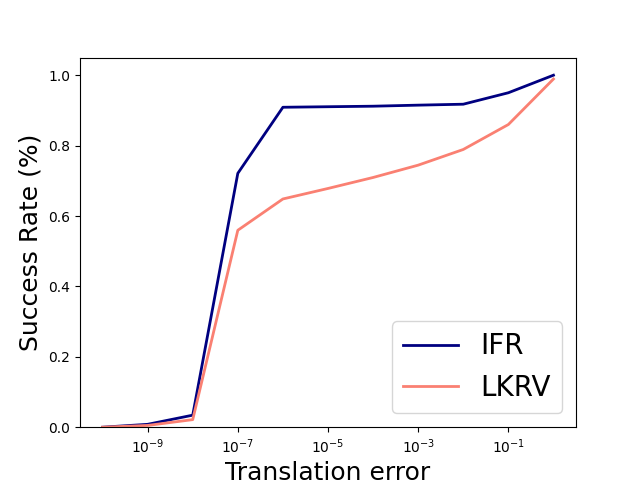}
		\caption{Translation threshold. }
	\end{subfigure}
	\caption{Success Rate. The $x$-axis is logarithmic.}
	\label{fig:modelnet_succ}
\end{figure}

To further demonstrate the competitive performance, we plot the success to error threshold curve in \cref{fig:modelnet_succ}.
Even at a very small maximum threshold, the success rates of our method and DL-based PointNetLK-Revisited are still similar and at a high level, well-reflecting the performance of our non-learning IFR model.

\subsection{Efficiency}
In \cite{li2021pointnetlk}, PointnetLK-Revisited shows a better efficiency than other baselines with a sample of $10^4$ points on a single CPU. It is feasible to test on CPU excluding the acceleration techniques as the time complexity is more intuitively reflected from the time cost. 
Similarly, we also sample $10^4$ points and test with our Intel Core i7-10710U CPU at 1.10GHz (a low power processor). PointNetLK-Revisited took 1.97s, while our model took 0.27s.
\subsection{Robustness Test}
Data collected with real sensors usually suffer from noise and sparsity.
Thus to further assess the robustness of our algorithm, we test for these properties.
PointnetLK-Re shows much better robustness than other baselines~\cite{li2021pointnetlk}.
Thus, we merely compare with the best PointnetLK-Re and shows scores with it.

\subsubsection{Different Density}

\begin{figure}[]
	\centering
	\begin{subfigure}[t]{.5\linewidth}
		\centering
		\includegraphics[width=1\linewidth]{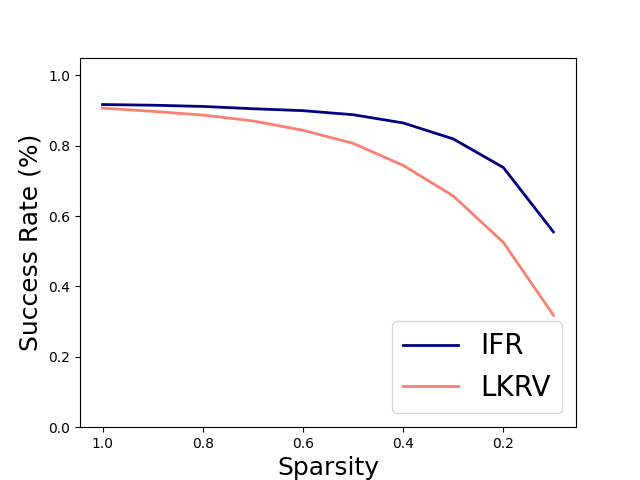}
		\caption{Sparse data.}
		\label{fig:sparsity}
	\end{subfigure}~
	\begin{subfigure}[t]{.5\linewidth}
		\centering
		\includegraphics[width=1\linewidth]{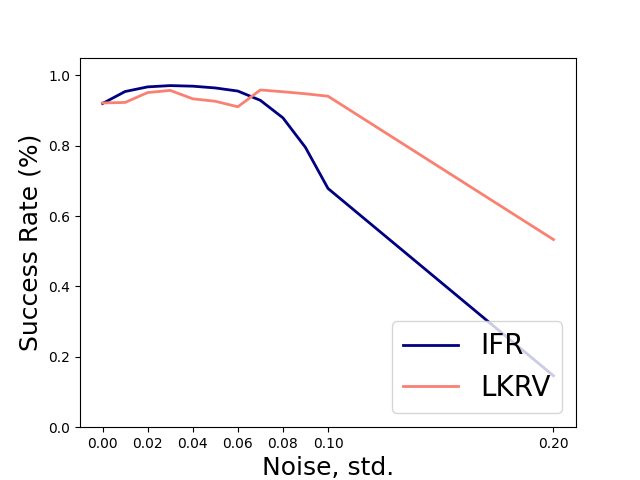}
		\caption{Robustness to noise.}
		\label{fig:noise}
	\end{subfigure}
	\caption{Robustness to noise and sparsity.}
\end{figure}

We implement the test on ModelNet Test set used in the previous part.
Due to the large requirement on GPU memory for PointNetLK-Re, we set the upper bound of point number to \num200,000.
Then we randomly select points from the sample point cloud.
The performance of our IFR algorithm is demonstrated in \cref{fig:sparsity}.
It outperforms the DL state of the art on the whole range of sparsity.

\subsubsection{Different Noise Level}

The noise robustness test is similar to the previous setting.
To add noise to the data, we randomly generate Gaussian Noises with
$\sigma \in \{0.01, 0.02\cdots 0.1, 0.2  \}$ to every point in both point clouds.
Our IFR uses the distance field as a basis.
Therefore, the noise is addressed at (1) data preprocessing, (2) smoothed distance field, and (3) on our general algorithm side.
To clarify the capability and limitation of our method, our algorithm deals with noise at level (3), i.e., on the algorithm side by taking $k=5$ nearest neighbors as the mean feature at the encoding step. 

From \cref{fig:noise}, we observe that in the range [\num{0}, \num{0.06}] meter in the noise, our IFR achieves a relative or even better success rate.
However, when the noise becomes extremely high, \SI{0.1}{\metre} or \SI{0.2}{\metre} for example, our curve drops faster than the DL-based PointNetLK-Re.
However, this extreme noise is not expected in real life, i.e., for realistic sensors. 
Therefore, in the next part, we will evaluate our IFR algorithm in real scenes.

\subsection{Real Indoor Data Test}
\label{sec:indoor_test}
After the robustness test, we turn to the real dataset that combines various aspects and challenges in the real world. 

For the synthetic tests, the distance fields are very similar and the generation within uniform cube or neighborhood will not affect the result.
However, in the real scene test, the distance fields for the two point cloud are usually partially different, which is more severe in space with large distances from the surface.
Thus points far from the surface will falsely affect the result.
Thus, we use points in the neighborhood for the real scene.

In the implementation, we draw $\V P_a$ near the point cloud $\V P_\tau$ with a Gaussian distribution centered at the surface points.
%In addition, truncation from~\cref{sec:trunc} is also applied.

The scores are given in \cref{tb:complex_generalizability}.
In the table, ICP achieves even better scores than any of the learning methods.
However, our model outperforms all.
With a truncation strategy (in \cref{sec:trunc}), the registration even achieves \num1.106 Median Rot. Error and \num0.049 Median Trans. Error. Ablation study in \cref{sec:ablation} is implemented to demonstrate the functionality of different modules.
We also outperform PointNetLK-Re but do not add it to this table.
The reason is detailed in supplementary$^2$.%~\cref{sec:why_no_}.

\begin{table}[]
  \centering
	\caption{Performance on complex, real-world scenes. All learning methods were trained on the synthetic ModelNet40 dataset and tested on the real-world 3DMatch dataset to investigate their generalizability across data with different distributions.}
\begin{threeparttable}
	\begin{adjustbox}{width=\columnwidth}
		\centering
		\begin{tabular}{lcccc}\toprule
			& \multicolumn{2}{c}{Rot. Error (degrees)} & \multicolumn{2}{c}{Trans. Error (m)} \\ \cmidrule(lr){2-3} \cmidrule(lr){4-5}
			Algorithm & RMSE~$\downarrow$ & Median~$\downarrow$ & RMSE~$\downarrow$ & Median~$\downarrow$ \\ \midrule
			ICP$^*$~\cite{besl1992method}   & 24.772 & 4.501 & 1.064 & \textbf{0.149} \\
			DCP$^*$~\cite{wang2019deep}   & 53.905 & 23.659 & 1.823 & 0.784 \\
			DeepGMR$^*$~\cite{yuan2020deepgmr} & 32.729 & 16.548 & 2.112 & 0.764 \\
			PointNetLK$^*$~\cite{aoki2019pointnetlk} & 28.894 & 7.596 & 1.098 & 0.260\\\midrule
			Ours & \textbf{23.814} & \textbf{4.191} & \textbf{1.057} & {0.161} \\ 
			Ours w/ trunc & 40.42 & \textbf{1.106} & 1.546 & \textbf{0.049} \\
			\bottomrule       
		\end{tabular}
	\end{adjustbox}
\begin{tablenotes}
	\item[$^*$] Values taken from \cite{li2021pointnetlk}.
\end{tablenotes}
\end{threeparttable}
	\vspace{0.01ex}
	\label{tb:complex_generalizability}
	\vspace{-.5cm}
\end{table}

\begin{table*}[]
  \centering
	\caption{RMSE on KITTI with artificial transformation. As compared deep learning based methods are trained on 00-07 sequences, we run tests on 08-10 sequences.}
	\begin{threeparttable}
		\begin{adjustbox}{width=1.9\columnwidth}
			\centering
			\begin{tabular}{lccccccc}\toprule
				&ICP Point2Point$^*$~\cite{besl1992method} & ICP Point2Plane$^*$~\cite{besl1992method} & G-ICP$^*$~\cite{segal2009generalized} & 3DFeat-Net$^*$~\cite{yew20183dfeat} & DeepVCP$^*$~\cite{lu2019deepvcp} & DeepCLR$^*$~\cite{horn2020deepclr} & Ours \\ \midrule
				Rot. Err. (deg.) & 0.088 & 0.079 & \textbf{0.029} & 0.199 & 0.164 & 0.053 & 0.088 \\ \midrule
				Trans. Err.& 0.177 & 0.140 & 0.109 & 0.116 & 0.071 & 0.080 & \textbf{0.068} \\
				\bottomrule
			\end{tabular}
		\end{adjustbox}
		\begin{tablenotes}
			\item[$^*$] Values taken from \cite{horn2020deepclr}.
		\end{tablenotes}
	\end{threeparttable}
	\vspace{0.01ex}
	
	\label{tb:kitti_acc}
\end{table*}

\subsection{Outdoor Data Test}
From our knowledge, the implicit function based reconstructions have only been applied to synthetic and indoor data sets, i.e., that is the basic case for use.
But to explore the capacity of our approach, we also test it on well-known outdoor LiDAR data, i.e., the KITTI-odometry.
Because the KITTI point cloud has varying density, in implementation we use an ISS-keypoint detector to sample points as surface points $\V P_{\tau,sub}$ ($\V P_{\tau,sub}$ for pseudo-set generation). From which, we generate $\V P_a$ as for the indoor test. The same truncation strategy as for the indoor test is also applied in this test.

%FIXED: Please spell out if space, i.e., err. -> error
The quantitative evaluation compared to the state-of-the-art (traditional and DL-based registration) on outdoor data is given in \cref{tb:kitti_acc}.
Our method achieves the best translation error and competitive rotation error to deep learning methods.
However, G-ICP scores best on the rotation error.

\begin{figure}[]

	\centering
	\includegraphics[width=1.\linewidth]{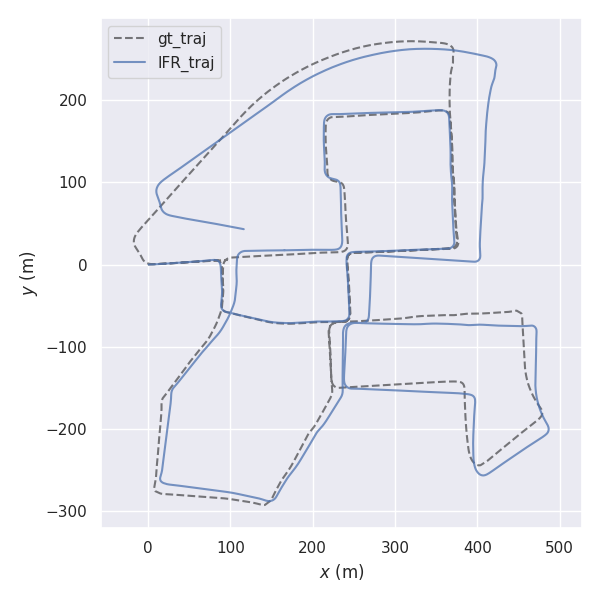}
	\caption{Registered trajectory for Kitti-odometry sequence 00 without loop-closure.}
	\label{fig:kitti_traj}
\end{figure}

In addition, we predict sequence 00 and plot it in \cref{fig:kitti_traj}.
Without any loop closure, it still provides a highly accurate trajectory.

%\subsubsection{KITTI Trajectory without Close-loop}
%\label{sec:kitti}

\subsection{Ablation Study}
\label{sec:ablation}
We implement an ablation study to show the impact of three different modules: IRLS, truncation, and pseudo-set generation. Because the different settings do not obviously improve the synthetic test, the study is applied to indoor test with the 3DMatch dataset. 

In \cref{tab:ablation}, ``without IRLS'' option disables IRLS and uses \cref{eq:solve_xi} to solve the transformation parameters $\xi$, ``with truncation'' option truncate dimensions following \cref{sec::setting}, uniform pseudo and neighborhood pseudo options are following the experiment setting at \cref{sec:synthetic_test} and \cref{sec:indoor_test} respectively. We plot two types of pseudo sets in supplementary material$^2$. % we plot  in \cref{sec:pseudo_demo}.

The full table with eight tests is provided at \cref{tab:ablation}.
With one option change, firstly, with IRLS or without IRLS, we observe that ``with IRLS'' group \#(5,6,7,8) shows a smaller median error than its corresponding test in group \#(1,2,3,4) respectively. This also holds for major ``with truncation'' group \#(4,7,8) to respectively exceed ``without truncation'' group \#(2,5,6). 
Moreover, major neighborhood pseudo group \#(4,6,8) shows a smaller median error than their corresponding uniform pseudo group \#(3,5,7) respectively.

The above comparison shows the effects of different modules (with IRLS, with truncation, and neighborhood pseudo). With all three modules equipped, \#8 scores the best.
%Using IRLS to solve transformation, \#8 shows better accuracy than direct solve in \#4. 
%With truncation to filter out some dimensions, \#8 also exceed \#6 that use all dimensions. 
%On the pseudo set generation, neighborhood pseudo option \#8 also fully exceed the \#7 with uniform pseudo. 

%FIXED: Are abreviations really needed?
However, with a truncation, our model (in~\cref{sec:trunc}) overfits two frames easier.
As shown in \cref{tab:ablation}, our model with truncation strategy achieves very accurate median rotatational and translational error. 
Though it provides more accurate results.
But its higher RMSE in \cref{tab:ablation} means that when it fails, it solves completely wrong.

\begin{table*}[htbp]
	\centering
	\caption{Ablation Study on IRLS, truncation and pseudo-set generation.}
	\label{tab:ablation}
	\resizebox{.8\textwidth}{!}{\begin{tabular}{c|cc|cc|cc|cccc}
			\hline
			&&&&&uniform&neighborhood& \multicolumn{2}{c}{Rot. Error}& \multicolumn{2}{c}{Trans. Error}\\
			
			\tabincell{c}{\#} & w/ IRLS & w/o IRLS & w/ trunc  &  w/o trunc &   pseudo &   pseudo &RMSE &Median &RMSE&Median\\
			
			\hline
			1& &\checkmark&&\checkmark&\checkmark&&18.732&7.073&0.969&0.311 \\
			2& &\checkmark&&\checkmark&&\checkmark&25.074&7.393&1.129&0.265\\
			3& &\checkmark&\checkmark&&\checkmark&&46.522&7.503&1.607&0.281 \\
			4& &\checkmark&\checkmark&&&\checkmark&37.896&2.998&1.492&0.118 \\
			5& \checkmark&&&\checkmark&\checkmark&&19.042&5.732&0.972&0.248 \\
			6& \checkmark&&&\checkmark&&\checkmark&23.018&4.443&1.057&0.157 \\
			7& \checkmark&&\checkmark&&\checkmark&&44.102&4.687&1.617&0.160 \\
			8& \checkmark&&\checkmark&&&\checkmark&32.886&\textbf{0.913}&1.148&\textbf{0.036} \\
			\hline
	\end{tabular}}
\end{table*}
%\section{Lessons learned}
%DONE The should be no sub section, if there is only one subsection
% Either add another subsubsection, or we remove the heading here
%\subsubsection{Limitations}

%Our work has limitations: (1) Our model suffers from large noise. Distance field is easily affected by noise as the filed value is determined by its nearest occupation but not density of occupation. One possible solution is to combined with regression based implicit functions (e.g. DL-based implicit function) that smooth the noise out. (2) 

\section{Conclusion and future work}
%\textbf{Limitations.} Our approach has limitations: (1) It is not robust to handle the outdoor registration for the complex context, sparse points, such as KITTI-odometr, see supplementary~\cref{sec:kitti}. 
In this paper, we have presented an indirect point cloud registration method using a pseudo third point set.
Our method registers two distance fields by only moving the pseudo-points, reducing the redundant movement of original point clouds.
In addition, we have extended the Feature Metric Registration class algorithms to a non-deep learning setting.
While using the analytical Jacobian, our pseudo-points based design highly reduces the theoretical space and time complexity.
From the comparison, our non-learning method achieved competitive and for some scenes better results than those learned models. 

%FIXED See above
%\subsection{Future Work}

%FIXED this needs to be finalized
In the future, we plan to embed this registration algorithm in an implicit-mapping based SLAM framework.
{\small
	\bibliographystyle{IEEEtran}
	\bibliography{egbib}

\begin{thebibliography}{10}
\providecommand{\url}[1]{#1}
\csname url@rmstyle\endcsname
\providecommand{\newblock}{\relax}
\providecommand{\bibinfo}[2]{#2}
\providecommand\BIBentrySTDinterwordspacing{\spaceskip=0pt\relax}
\providecommand\BIBentryALTinterwordstretchfactor{4}
\providecommand\BIBentryALTinterwordspacing{\spaceskip=\fontdimen2\font plus
\BIBentryALTinterwordstretchfactor\fontdimen3\font minus
  \fontdimen4\font\relax}
\providecommand\BIBforeignlanguage[2]{{%
\expandafter\ifx\csname l@#1\endcsname\relax
\typeout{** WARNING: IEEEtran.bst: No hyphenation pattern has been}%
\typeout{** loaded for the language `#1'. Using the pattern for}%
\typeout{** the default language instead.}%
\else
\language=\csname l@#1\endcsname
\fi
#2}}

\bibitem{koch2016multi}
P.~Koch, S.~May, M.~Schmidpeter, M.~K{\"u}hn, C.~Pfitzner, C.~Merkl, R.~Koch,
  M.~Fees, J.~Martin, D.~Ammon, \emph{et~al.}, ``Multi-robot localization and
  mapping based on signed distance functions,'' \emph{Journal of Intelligent \&
  Robotic Systems}, vol.~83, no.~3, pp. 409--428, 2016.

\bibitem{yuan2021self}
Y.~Yuan, D.~Borrmann, J.~Hou, Y.~Ma, A.~N{\"u}chter, and S.~Schwertfeger,
  ``Self-supervised point set local descriptors for point cloud registration,''
  \emph{Sensors}, vol.~21, no.~2, p. 486, 2021.

\bibitem{huang2021di}
J.~Huang, S.-S. Huang, H.~Song, and S.-M. Hu, ``Di-fusion: Online implicit 3d
  reconstruction with deep priors,'' in \emph{Proceedings of the IEEE/CVF Conf.
  on Computer Vision and Pattern Recognition}, 2021, pp. 8932--8941.

\bibitem{slavcheva2018signed}
M.~Slavcheva, ``Signed distance fields for rigid and deformable 3d
  reconstruction,'' Ph.D. dissertation, Technische Universit{\"a}t M{\"u}nchen,
  2018.

\bibitem{li2019usip}
J.~Li and G.~H. Lee, ``Usip: Unsupervised stable interest point detection from
  3d point clouds,'' in \emph{Proceedings of the IEEE/CVF International Conf.
  on Computer Vision}, 2019, pp. 361--370.

\bibitem{yew20183dfeat}
Z.~J. Yew and G.~H. Lee, ``3dfeat-net: Weakly supervised local 3d features for
  point cloud registration,'' in \emph{Proceedings of the Europ. Conf. on
  Computer Vision (ECCV)}, 2018, pp. 607--623.

\bibitem{lu2020rskdd}
F.~Lu, G.~Chen, Y.~Liu, Z.~Qu, and A.~Knoll, ``Rskdd-net: Random sample-based
  keypoint detector and descriptor,'' \emph{arXiv preprint arXiv:2010.12394},
  2020.

\bibitem{bai2020d3feat}
X.~Bai, Z.~Luo, L.~Zhou, H.~Fu, L.~Quan, and C.-L. Tai, ``D3feat: Joint
  learning of dense detection and description of 3d local features,'' in
  \emph{Proceedings of the IEEE/CVF Conf. on Computer Vision and Pattern
  Recognition}, 2020, pp. 6359--6367.

\bibitem{yang2020teaser}
H.~Yang, J.~Shi, and L.~Carlone, ``Teaser: Fast and certifiable point cloud
  registration,'' \emph{IEEE Transactions on Robotics}, vol.~37, no.~2, pp.
  314--333, 2020.

\bibitem{choy2020deep}
C.~Choy, W.~Dong, and V.~Koltun, ``Deep global registration,'' in
  \emph{Proceedings of the IEEE/CVF conference on computer vision and pattern
  recognition}, 2020, pp. 2514--2523.

\bibitem{lu2021hregnet}
F.~Lu, G.~Chen, Y.~Liu, L.~Zhang, S.~Qu, S.~Liu, and R.~Gu, ``Hregnet: A
  hierarchical network for large-scale outdoor lidar point cloud
  registration,'' in \emph{Proceedings of the IEEE/CVF International Conf. on
  Computer Vision}, 2021, pp. 16\,014--16\,023.

\bibitem{aoki2019pointnetlk}
Y.~Aoki, H.~Goforth, R.~A. Srivatsan, and S.~Lucey, ``Pointnetlk: Robust \&
  efficient point cloud registration using pointnet,'' in \emph{Proceedings of
  the IEEE/CVF Conf. on Computer Vision and Pattern Recognition}, 2019, pp.
  7163--7172.

\bibitem{huang2020feature}
X.~Huang, G.~Mei, and J.~Zhang, ``Feature-metric registration: A fast
  semi-supervised approach for robust point cloud registration without
  correspondences,'' in \emph{Proceedings of the IEEE/CVF Conf. on Computer
  Vision and Pattern Recognition}, 2020, pp. 11\,366--11\,374.

\bibitem{li2021pointnetlk}
X.~Li, J.~K. Pontes, and S.~Lucey, ``Pointnetlk revisited,'' in
  \emph{Proceedings of the IEEE/CVF Conf. on Computer Vision and Pattern
  Recognition}, 2021, pp. 12\,763--12\,772.

\bibitem{slavcheva2016sdf}
M.~Slavcheva, W.~Kehl, N.~Navab, and S.~Ilic, ``Sdf-2-sdf: Highly accurate 3d
  object reconstruction,'' in \emph{Europ. Conf. on Computer Vision}.\hskip 1em
  plus 0.5em minus 0.4em\relax Springer, 2016, pp. 680--696.

\bibitem{slavcheva2016sdf2}
M.~Slavcheva and S.~Ilic, ``Sdf-tar: Parallel tracking and refinement in rgb-d
  data using volumetric registration.'' in \emph{BMVC}, 2016.

\bibitem{yuan2019incrementally}
Y.~Yuan and S.~Schwertfeger, ``Incrementally building topology graphs via
  distance maps,'' in \emph{2019 IEEE International Conf. on Real-time
  Computing and Robotics (RCAR)}.\hskip 1em plus 0.5em minus 0.4em\relax IEEE,
  2019, pp. 468--474.

\bibitem{gentle2007matrix}
J.~E. Gentle, ``Matrix algebra,'' \emph{Springer texts in statistics, Springer,
  New York, NY, doi}, vol.~10, pp. 978--0, 2007.

\bibitem{besl1992method}
P.~J. Besl and N.~D. McKay, ``Method for registration of 3-d shapes,'' in
  \emph{Sensor fusion IV: control paradigms and data structures}, vol.
  1611.\hskip 1em plus 0.5em minus 0.4em\relax International Society for Optics
  and Photonics, 1992, pp. 586--606.

\bibitem{wang2019deep}
Y.~Wang and J.~M. Solomon, ``Deep closest point: Learning representations for
  point cloud registration,'' in \emph{Proceedings of the IEEE/CVF
  International Conf. on Computer Vision}, 2019, pp. 3523--3532.

\bibitem{yuan2020deepgmr}
W.~Yuan, B.~Eckart, K.~Kim, V.~Jampani, D.~Fox, and J.~Kautz, ``Deepgmr:
  Learning latent gaussian mixture models for registration,'' in \emph{Europ.
  Conf. on Computer Vision}.\hskip 1em plus 0.5em minus 0.4em\relax Springer,
  2020, pp. 733--750.

\bibitem{park2019deepsdf}
J.~J. Park, P.~Florence, J.~Straub, R.~Newcombe, and S.~Lovegrove, ``Deepsdf:
  Learning continuous signed distance functions for shape representation,'' in
  \emph{Proceedings of the IEEE/CVF Conf. on Computer Vision and Pattern
  Recognition}, 2019, pp. 165--174.

\bibitem{chibane2020neural}
J.~Chibane, A.~Mir, and G.~Pons-Moll, ``Neural unsigned distance fields for
  implicit function learning,'' \emph{arXiv preprint arXiv:2010.13938}, 2020.

\bibitem{wu20153d}
Z.~Wu, S.~Song, A.~Khosla, F.~Yu, L.~Zhang, X.~Tang, and J.~Xiao, ``3d
  shapenets: A deep representation for volumetric shapes,'' in
  \emph{Proceedings of the IEEE conference on computer vision and pattern
  recognition}, 2015, pp. 1912--1920.

\bibitem{chang2015shapenet}
A.~X. Chang, T.~Funkhouser, L.~Guibas, P.~Hanrahan, Q.~Huang, Z.~Li,
  S.~Savarese, M.~Savva, S.~Song, H.~Su, \emph{et~al.}, ``Shapenet: An
  information-rich 3d model repository,'' \emph{arXiv preprint
  arXiv:1512.03012}, 2015.

\bibitem{zeng20173dmatch}
A.~Zeng, S.~Song, M.~Nie{\ss}ner, M.~Fisher, J.~Xiao, and T.~Funkhouser,
  ``3dmatch: Learning local geometric descriptors from rgb-d reconstructions,''
  in \emph{Proceedings of the IEEE conference on computer vision and pattern
  recognition}, 2017, pp. 1802--1811.

\bibitem{geiger2012we}
A.~Geiger, P.~Lenz, and R.~Urtasun, ``Are we ready for autonomous driving? the
  kitti vision benchmark suite,'' in \emph{2012 IEEE conference on computer
  vision and pattern recognition}.\hskip 1em plus 0.5em minus 0.4em\relax IEEE,
  2012, pp. 3354--3361.

\bibitem{horn2020deepclr}
M.~Horn, N.~Engel, V.~Belagiannis, M.~Buchholz, and K.~Dietmayer, ``Deepclr:
  Correspondence-less architecture for deep end-to-end point cloud
  registration,'' in \emph{2020 IEEE 23rd International Conf. on Intelligent
  Transportation Systems (ITSC)}.\hskip 1em plus 0.5em minus 0.4em\relax IEEE,
  2020, pp. 1--7.

\bibitem{segal2009generalized}
A.~Segal, D.~Haehnel, and S.~Thrun, ``Generalized-icp.'' in \emph{Robotics:
  science and systems}, vol.~2, no.~4.\hskip 1em plus 0.5em minus 0.4em\relax
  Seattle, WA, 2009, p. 435.

\bibitem{lu2019deepvcp}
W.~Lu, G.~Wan, Y.~Zhou, X.~Fu, P.~Yuan, and S.~Song, ``Deepvcp: An end-to-end
  deep neural network for point cloud registration,'' in \emph{Proceedings of
  the IEEE/CVF International Conf. on Computer Vision}, 2019, pp. 12--21.

\end{thebibliography}
}

\end{document}